\documentclass[letterpaper, 10 pt, conference]{ieeeconf}  %

\usepackage[utf8]{inputenc}
\usepackage{booktabs}
\usepackage{siunitx}

\usepackage{hyperref}
\usepackage[caption=false]{subfig}  %

\usepackage{balance}

\usepackage[export]{adjustbox}
\usepackage{floatrow}
\newfloatcommand{capbtabbox}{table}[][\FBwidth]

\newcommand{\myparagraph}[1]{\textbf{#1}~} %

\newcommand{\mytimes}{\medmuskip=0mu\times}
\newcommand{\meters}{\textrm{m}}
\newcommand{\dense}{\thickmuskip=2mu}
\DeclareRobustCommand*\pct{\scalebox{.9}{\%}}  %

\def\abovestrut#1{\rule[0in]{0in}{#1}\ignorespaces}
\newcommand{\smallgap}{\abovestrut{0.15in}}

\newcommand{\eqref}[1]{(\ref{#1})}
\newcommand{\figref}[1]{Fig.~\ref{#1}}

\newcommand{\tabref}[1]{Table~\ref{#1}}

\newcommand{\ie}{i.e.}
\newcommand{\eg}{e.g.}

\newcommand{\yaw}{\ensuremath{\psi}}

\newcommand{\conv}[4]{#1-#2-#3-#4}
\newcommand{\fc}[2]{#1-#2}

\newcommand\blfootnote[1]{%
  \begingroup
  \renewcommand\thefootnote{}\footnote{#1}%
  \addtocounter{footnote}{-1}%
  \endgroup
}

\usepackage[disable]{todonotes}

\newcommand{\best}[1]{\textbf{#1}}

\newcommand{\workshop}[1]{}

\IEEEoverridecommandlockouts                              %

\title{\LARGE \bf
Differentiable Mapping Networks: Learning Structured Map Representations for Sparse Visual Localization
}

\author{Peter Karkus$^{1,2}$, Anelia Angelova$^{1}$, Vincent Vanhoucke$^{1}$, and Rico Jonschkowski$^{1}$%
\thanks{$^{1}$Robotics at Google {\tt\footnotesize [anelia,vanhoucke,rjon]@google.com}}%
\thanks{$^{2}$National University of Singapore {\tt\footnotesize karkus@comp.nus.edu.sg}}%
}

\begin{document}

\maketitle
\thispagestyle{empty}
\pagestyle{empty}

\begin{abstract}
Mapping and localization, preferably from a small number of observations, are fundamental tasks in robotics. We address these tasks by combining spatial structure (differentiable mapping) and end-to-end learning in a novel neural network architecture: the Differentiable Mapping Network (DMN). The DMN constructs a spatially structured view-embedding map and uses it for subsequent visual localization with a particle filter. Since the DMN architecture is \emph{end-to-end differentiable}, we can jointly learn the map representation and localization using gradient descent. We apply the DMN to sparse visual localization, where a robot needs to localize in a new environment with respect to a small number of images from known viewpoints. We evaluate the DMN using simulated environments and a challenging real-world Street View dataset. We find that the DMN learns effective map representations for visual localization. The benefit of spatial structure increases with larger environments, more viewpoints for mapping, and when training data is scarce.
Project website: \url{https://sites.google.com/view/differentiable-mapping}.
\end{abstract}

\section{Introduction}

To efficiently operate in a new environment, robots must be able to build a \emph{map} -- an internal representation of the environment -- even from few observations. 
But how should that map be represented and what information should be stored in it to enable downstream tasks, e.g., localization?
Classic approaches use a fixed map representation with strong spatial structure, such as voxels or point clouds. This makes them applicable to a wide range of robotic tasks. In contrast, data-driven deep neural networks can learn rich representations by optimizing them directly for a downstream task. Eslami et al.~\cite{eslami2018neural}, for example, learn to construct environment representations from a few images with known viewpoints, to then generate images from novel viewpoints. The challenge for learning is choosing suitable \emph{priors} that enable generalization. %
Spatial structure can serve as such priors, and allow us to combine the best of both worlds, incorporating assumptions from classical approaches while leveraging the power of deep neural networks to learn a flexible and effective map representation for the downstream task.

\begin{figure}[t!]
    \centering
    \includegraphics[width=0.9\textwidth]{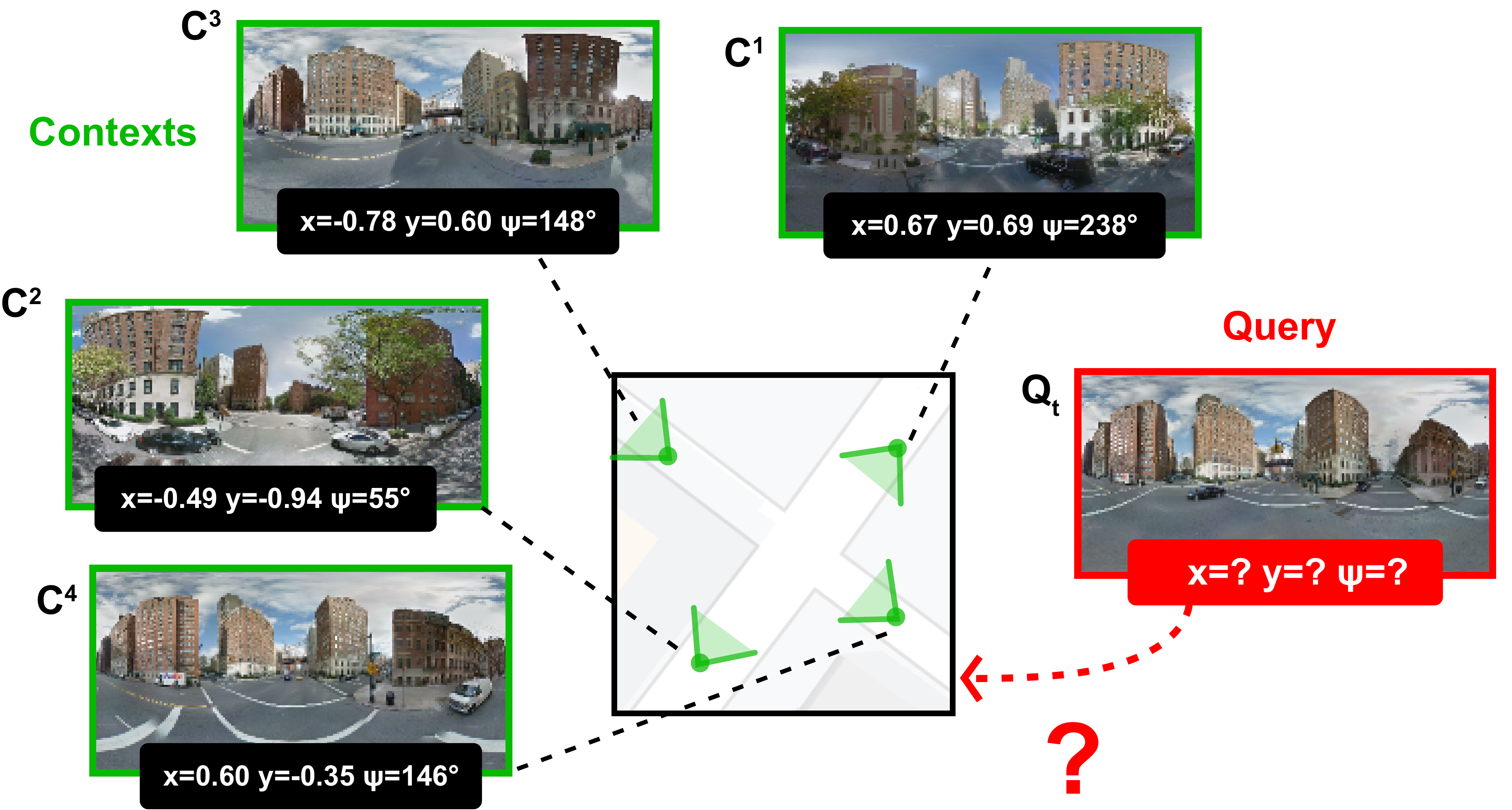}
    \caption{Sparse visual localization in the Street View domain. The robot receives 4 context image-pose pairs, $C^1, C^2, C^3, C^4$, and a query image $Q_1$; and it needs to estimate the pose of the query image.}
 \label{fig:localization}
\end{figure}

In this paper we explore how structure and learning can be combined in the context of a sparse visual localization task~(\figref{fig:localization}). In this task
a robot has access to a handful of visual observations from known viewpoints to build a map, and then it needs to localize with respect to this map given a new sequence of visual observations. The task is challenging due to the small number of observations in which the relevant spatial information is encoded in rich visual features. While this setting is not widely studied, with a notable exception~\cite{rosenbaum2018learning}, it has practical importance for high impact real-world applications, such as i) localizing a vehicle anywhere on Earth using only an onboard camera and pre-collected visual data (e.g., Street View~\cite{mirowski2018learning}), ii) mapping for multi-robot cooperative tasks with limited communication bandwidth, e.g., for search and rescue, and iii) localization in warehouses with frequently changing visual appearance.

\begin{figure}[!t]
\begin{floatrow}
\ffigbox[.85\textwidth]{%
\vspace{.1cm}%
\includegraphics[width=0.85\textwidth,left]{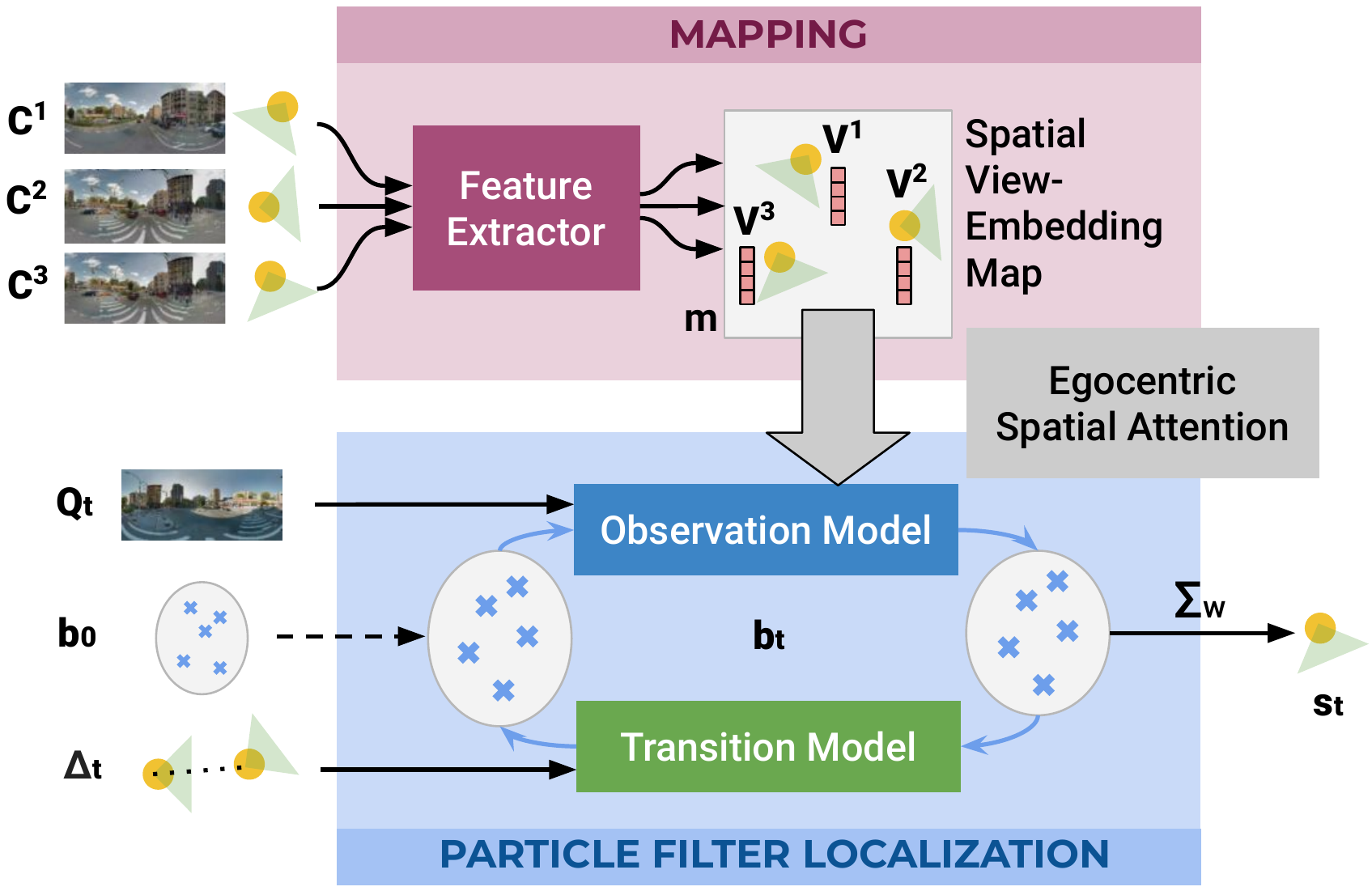}%
}{%
\caption{The Differentiable Mapping Network (DMN)  schematic.}%
\label{fig:dmn_network}}%
\hspace{.08\textwidth}
\end{floatrow}
\end{figure}

To learn map representations for this downstream task, we introduce the \emph{Differentiable Mapping Network}~(DMN), a novel neural network architecture with spatial structure as a prior for mapping (\figref{fig:dmn_network}).
Given a set of image-pose pairs, the DMN constructs a structured latent map representation consisting of pairs of viewpoint coordinates and learned image embeddings. Based on this map the model performs visual localization with a differentiable particle filter~\cite{jonschkowski2018differentiable,karkus2018particle}. 
During localization, the map is transformed into the coordinate frame of each particle, which provides the model with a 
simple but powerful prior for spatial reasoning.

We evaluate our method both in synthetic 3D environments~\cite{eslami2018neural} and on real-world Street View data~\cite{mirowski2018learning}. During training, we sample different environments, let the model build a map from context image-pose pairs, run particle filter localization for query images, and optimize the network based on the difference of predicted and true query poses. 
During inference, we test the model in previously unseen environments. Our results show that the DMN is effective in a variety of sparse localization settings. High performance critically depends on the structured map representation and the advantage
increases with less training data, larger environments, and more context observations for mapping.

\section{Related work}

Classical approaches to mapping and localization rely on fixed, structured map representations such as grids \cite{thrun2003learning}, point clouds \cite{henry2012rgb}, feature points \cite{mur2015orb}, or  landmarks \cite{montemerlo2002fastslam}. 
With the emergence of data-driven methods, especially deep learning, there has been increasing interest in learning to represent the map in a neural memory \cite{kendall2015posenet, zamir2016generic, zhang2017neural, eslami2018neural, rosenbaum2018learning}. 
Since generic neural networks require large amounts of training data \cite{eslami2018neural}, encoding spatial structure in the memory is a promising direction that has been explored in the context of scene representation and mapping  \cite{henriques2018mapnet, gupta2017cognitive, huynh2019multigrid, tung2019learning}, localization \cite{radwan2018vlocnet++, parisotto2018global}, and SLAM~\cite{henriques2018mapnet, parisotto2018neural, avraham2019empnet}. %
A neural network can even incorporate structure by encoding modules of \emph{differentiable algorithms}~\cite{karkus2019differentiable}. Examples include robot localization \cite{jonschkowski2018differentiable, karkus2018particle}, planning \cite{tamar2016value, oh2017value, farquhar2017treeqn, guez2018learning}, and control \cite{ okada2017path, amos2018differentiable, pereira2018mpc}.

Our work is aligned with these methods, but focuses on differentiable mapping that learns a latent map optimized for a downstream task (in our case sparse localization). Instead of using the map to generate observations as in~\cite{rosenbaum2018learning}, we use it discriminatively to evaluate candidate poses for localization: when querying the map with a query image and coordinate, viewpoints are transformed into the query coordinate frame and embeddings are combined with the query image features through a spatial attention mechanism. Compared to unstructured maps in prior work~\cite{eslami2018neural, rosenbaum2018learning}, our structured map avoids lossy aggregation by maintaining embeddings for each context input, it 
allows for generalization from limited training data, and the differentiable architecture allows for powerful task-oriented learning.  Context embeddings in a more general setting have been explored in~\cite{kim2019attentive}.

\section{Sparse Visual Localization} 
\label{sec:sparse_localization}

In the sparse visual localization task (\figref{fig:localization}) a robot receives a handful of RGB-image and pose pairs from a previously unseen environment, which we will refer to as \emph{context}, $C^i$, $\dense i=1:N_c$. From this sparse context the robot constructs a \emph{map}, $m$, and uses the map for subsequent localization, i.e., to estimate the pose $s_t$ for a sequence of \emph{query} RGB-images from novel viewpoints, $Q_t$. The robot is given the relative egomotion between time steps, $\Delta_t$, and an initial belief, $b_0$, that is potentially uninformed uniform. Query images are taken in the same environment as the context, but potentially at a much later time, which can be challenging in dynamic environments such as city streets.

\section{Differentiable Mapping Network}

\begin{figure}[!t]
\vspace{.1cm}%
\begin{floatrow}
\hspace{-.16\textwidth}
\ffigbox[.53\textwidth]{%
\hspace{-.04\textwidth}%
\includegraphics[width=.55\textwidth]{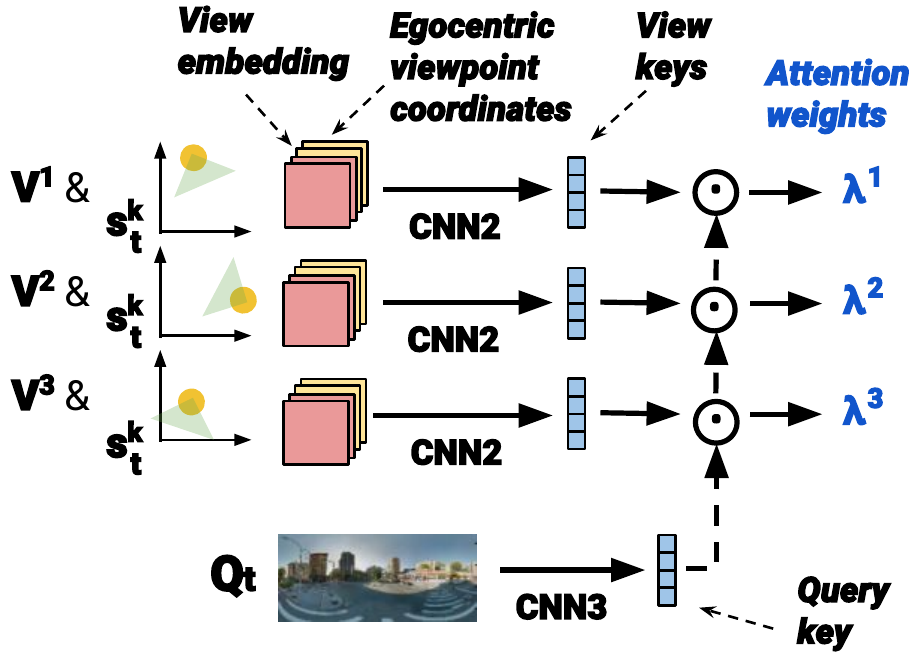}
}{\caption{Egocentric spatial attention.%
} \label{fig:attention}
}
\ffigbox[.56\textwidth]{%
\hspace{-.032\textwidth}%
\includegraphics[width=.58\textwidth]{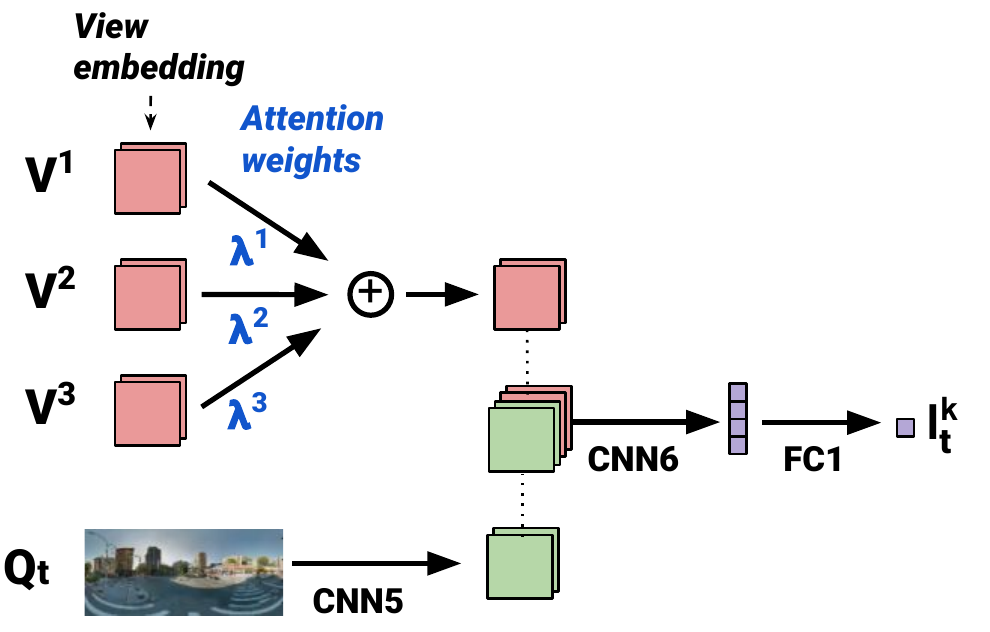}
}{\caption{Observation model.
}\label{fig:obs_model}
}
\end{floatrow}
\end{figure}

\myparagraph{The DMN Architecture.}
We introduce the DMN, a novel neural network architecture that learns a spatial view-embedding map for sparse localization (\figref{fig:dmn_network}). 
For each context $C^i$ the network computes an embedding $V^i$ (top left of \figref{fig:dmn_network}). The set of embeddings and viewpoint coordinates taken together make up the map representation $m$ (top), which is used for localization (bottom) using a \emph{differentiable particle filter}~\cite{jonschkowski2018differentiable,karkus2018particle}, where particles $s_t^k$ correspond to query pose candidates. Starting with a set of particles sampled from the initial belief $b_0$, the observation model updates the particle weights by comparing the query image $Q_t$ to the map $m$ using 
\emph{egocentric spatial attention}, i.e., attention based on the relative pose of the context viewpoints and the particle. The transition model updates particles with the egomotion input $\Delta_t$. At each time step the query pose output $s_t$ is estimated by the weighted mean of particles (via $\Sigma_w$). For simplicity we refer to the full approach of building a map and its use in particle filter localization as DMN. 

\myparagraph{Mapping.} %
The DMN builds a spatial view-embedding map, $\dense m = \langle V^i, s^i \rangle_{i=1:N_c}$, a set of view-embedding $V_i$ and viewpoint $s^i$ pairs. View-embeddings $V^i$ are features extracted from the corresponding context image $C^i$ with a CNN, denoted as CNN1 {\footnotesize(\conv{32}{2}{2}{relu}, \conv{32}{3}{1}{relu}, \conv{64}{2}{2}{relu}, \conv{64}{3}{1}{relu})}, where the notation is {\footnotesize (\conv{filters}{kernel}{strides}{activation})}. Weights are shared across contexts. $V^i$ has a shape of $\dense H \mytimes W \mytimes {D}$, where $H$ and $W$ are $1/4$ of the context image height and width, and $\dense {D}=64$. Viewpoints are planar poses $\dense s^i = (x, y, \sin{\yaw}, \cos{\yaw})$, where $\yaw$ is the yaw.

\myparagraph{Egocentric spatial attention.}
To ``read'' the map from a query viewpoint $s_t^k$ we introduce an \emph{egocentric spatial attention} mechanism (\figref{fig:attention}). The attention mechanism computes a weighted sum of view-embeddings using the scalar product of a \emph{query key} and a \emph{view key} as the weight, similarly to standard attention~\cite{vaswani2017attention}. Our attention is \emph{spatial}: it extracts view keys from view-embeddings concatenated with viewpoint coordinates. The attention is also \emph{egocentric}: viewpoint coordinates are transformed into an egocentric coordinate frame where the query viewpoint $s_t^k$ is the origin. This attention mechanism leverages the spatial structure of the map, which is expected to substantially reduce the difficulty of extracting useful features.

In our implementation viewpoints are transformed into the egocentric coordinate frame with simple geometric transformation. Coordinates are tiled to $\dense H \mytimes W \mytimes 4$, concatenated with the view-embedding $V^i$ along the last axis,
and passed through CNN2 {\footnotesize(\conv{64}{3}{1}{relu}, \conv{32}{3}{1}{lin})}, which gives a view key. The query key is extracted from the query image $Q_t$ with CNN3 {\footnotesize (\conv{32}{2}{2}{relu}, \conv{32}{3}{1}{relu}, \conv{64}{2}{2}{relu}, \conv{32}{3}{1}{relu}, \conv{64}{3}{1}{relu}, \conv{32}{3}{1}{lin})}. We take the scalar product of each view key and the query key %
to compute the attention weights $\lambda^i$. We further process view-embeddings with CNN4 {\footnotesize(\conv{128}{3}{1}{relu}, \conv{64}{3}{1}{relu})}, concatenate with the view keys 
(not shown in figures),
and sum them, weighting by the attention weights $\lambda^i$ (top left of \figref{fig:obs_model}). %

\begin{figure}[t]
    \centering
    \vspace{.1cm}%
    \includegraphics[width=0.99\textwidth]{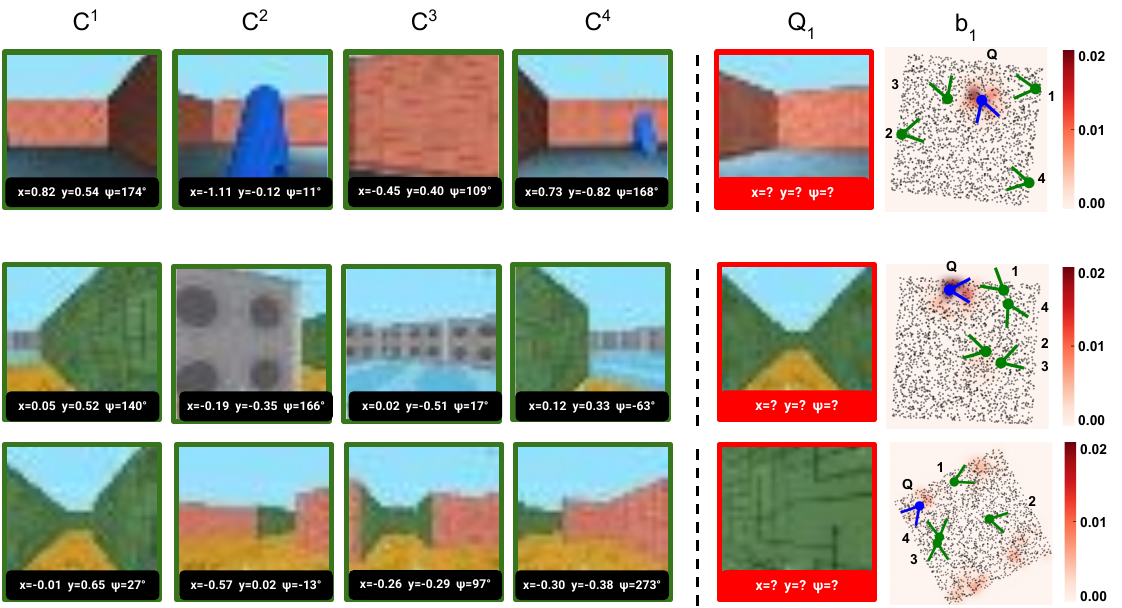}
    \caption{Examples for one-step global localization in the Rooms and Mazes domains. 
    The last column shows DMN particles (black), context viewpoints (green) and the unknown query pose (blue). Particle weights are visualized as a heat map aggregated over all yaw values.}
 \label{fig:gqn_examples}
\end{figure}

\myparagraph{Particle filter localization.}
Based on the map, the DMN performs subsequent sequential localization with a differentiable PF. The PF represents the belief as a set of weighted particles, $\dense b_t(s) \approx \langle s_t^k, \log{w_t^k} \rangle_{k=1:K}$,
where a particle $s_t^k$ is a candidate pose of the robot, i.e., the unknown query viewpoint. %
In the first step particles are sampled from the input initial belief $b_0$.
In each following step, particles are updated given new observations and the latent map. Particle poses are updated with a fixed transition model, $s_t^k = f_T(s_{t-1}^k, \Delta_t)$, using simple geometry and the egomotion input $\Delta_t$. Particle weights are updated with a learned observation model, $\log{w_t^k} = \log{l_\theta(Q_t, s_t^k, m)} + \log{w_{t-1}^k} + \eta$, given the query image $Q_t$, the particle pose $s_t^k$, the latent map $m$ and a normalization factor $\eta=-\log{\sum_{j=1:K}{e^{\log{w_t^j}}}}$. 

The observation model connects the map with the particle filter (\figref{fig:obs_model}): it estimates the conditional log-probability of observing $Q_t$, given a pose $s_t^k$ and the map $m$, that is, $l_\theta(Q_t, s_t^k, m) \approx \log{p(Q_t | s_t^k, m)}$. Our observation model is discriminative: it takes in $Q_t$, $s_t^k$, and $m$, and outputs a real value, $l_t^k$, a direct estimate of the unnormalized particle log-likelihood. The model first performs egocentric spatial attention over the view-embeddings of $m$ given the query image $Q_t$ and $s_t^k$, a particle treated as the query pose. The attention mechanism outputs map features that are concatenated with another set of features extracted from the query image with CNN5 {\footnotesize(\conv{32}{2}{2}{relu}, \conv{32}{3}{1}{relu}, \conv{64}{2}{2}{relu}, \conv{64}{3}{1}{relu}, \conv{128}{3}{1}{relu}, \conv{64}{3}{1}{relu})}). 
The first three layers of CNN4 and CNN5 are shared. The concatenated features are passed through another convolutional--dense component, CNN6 {\footnotesize(\conv{64}{3}{1}{relu}, \conv{64}{3}{1}{relu}, \conv{64}{3}{1}{relu}, flatten)} and FC1 {\footnotesize (\fc{512}{relu}, \fc{1}{relu})}. The output is a single value, the log-likelihood $l_t^k$. Log-likelihoods are estimated for each particle this way (learnable parameters are shared), normalized across particles with $\eta$, and used to update particle weights.

The output pose estimate is given by the mean particle weighted by the particle weights,  
$s_t = \sum_k{w_t^k s_t^k}$.  
Alternatively, a weighted kernel density estimate, e.g., a mixture of Gaussians, could be also used~\cite{jonschkowski2018differentiable}.
Note that $s_t$ depends on all inputs ($C^i, Q_t, \Delta_t, b_0)$ through the learned feature extractors of the mapping and the localization components. %

\myparagraph{End-to-end training.}
Since the DMN is end-to-end differentiable, we can train it to optimize mapping for the task of localization. 
To learn mapping this way, we sample different environments, different contexts and queries per environment, present them to the network, and optimize all parameters for localization by backpropagating gradients through the particle filter. The training loss is the mean-squared-error (MSE) between the pose estimate $s$ and true query pose $s^*$:
$\mathcal{L} = ||s-s^*||^2 = (x-x^*)^2 + (y-y^*)^2 + \alpha (\yaw-\yaw^*)^2$, 
where $\dense \alpha=0.5$ is a hyperparameter.
Importantly, we can vary both the number of contexts and the number of particles of a trained DMN because of the shared weights.

\begin{figure}[t]
\vspace{.15cm}%
\centering
    \includegraphics[width=0.999\textwidth]{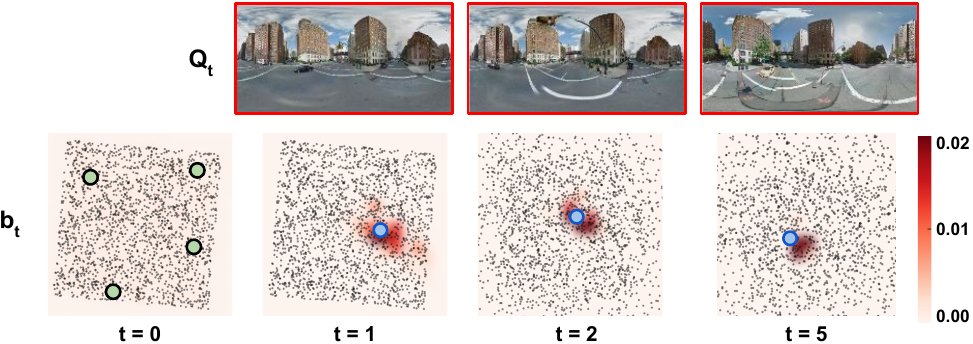}
    \caption{Sequential global localization example in the Street View domain.  Corresponding context images are shown in \figref{fig:localization}.
    The bottom row shows DMN particles using the same semantics as in \figref{fig:gqn_examples}.}
 \label{fig:sv_example}
\end{figure}

\section{Experiments}
\label{sec:experiments}

\subsection{Domains and datasets}
\label{sec:domains}

\myparagraph{3D simulated environments.} 
We first evaluate our approach in the established 3D simulated datasets, \emph{Rooms} and \emph{Mazes}~\cite{eslami2018neural,rosenbaum2018learning}. Examples are shown in \figref{fig:gqn_examples}.
Environments in the Rooms domain are square rooms with randomly generated objects and textures. The Maze environments are large mazes with randomly generated layouts and different textures. Here highly ambiguous observations create a challenge for localization. For the sparse localization task we randomly sample context image-pose pairs and query images-pose pairs from a given environment, and generate trajectories by randomly ordering the sampled query poses. We use 1\% of the original training set (100k environments and 10 images per environment for Rooms, 960 environments and 300 images per environment for Mazes), and a separate set of 5000 environments for evaluation. RGB images are resized to $32\mytimes32$. We treat each environment as a $\dense 40\meters\mytimes 40\meters$ area, 
but when entering the model, pose coordinates are normalized to $\dense [-1, 1]$.

\myparagraph{Street View dataset.}
We also perform experiments in the real-world Street View domain, using the recently released StreetLearn dataset~\cite{mirowski2018learning} which 
consists of panorama images with known viewpoints taken in New York City's Manhattan (\figref{fig:localization}). Street View
data has been used in prior work, \eg~\cite{zamir2014image, zamir2016generic, arandjelovic2016netvlad}, but we are not aware of work on mapping or sparse localization with the StreetLearn dataset. Note that sparse localization in this domain is particularly challenging because the environment can differ significantly between the times of mapping and localization. 
The data is collected such that cars and pedestrians typically move between consecutive images. Moreover, images from an area might be collected at different times, potentially years apart, during which seasons may change and new buildings may be built, etc. 
We split the dataset at the longitude of $-73.9857$ into a training set of 38,746 data points west and a test set of 16,359 data points east of the split.
For sparse localization we randomly sample areas of size $\dense 40\meters\mytimes40\meters$ with at least 10 image-pose pairs. 
Panorama RGB images are resized to $104\mytimes208$. Pose coordinates are normalized to $\dense [-1,1]$.

\begin{table}[t]
\caption{Global localization results for 
simulated environments%
}
\label{tab:main_joint}
\begin{center}
\scalebox{.73}{
\begin{tabular}{llcccccc} 
\toprule
\multicolumn{2}{c}{Approach} & 
\multicolumn{2}{c}{Sequential loc. ($\uparrow$)} &
\hskip 4em &
\multicolumn{2}{c}{One-step loc. ($\uparrow$)} %
\\
Map & Algorithm 
& Rooms & Mazes  & 
& Rooms & Mazes \\ 
\midrule

View-embed.  & DMN-PF (ours)  
& $95.6\pct$    & \best{75.9\pct}   &
& \best{93.5\pct{}}   & \best{27.0\pct}   
\\

View-embed.  &  Regr. (ours)           
& \best{97.2\pct}      & 39.9\pct{}      &
& $60.5\pct$      & $19.0\pct$       
\\
\hline

\smallgap

Latent image & PF                
& $94.0\pct$      & $69.4\pct$       &
& $90.5\pct$      & $21.1\pct$      
\\

Latent image & Regression  
& $9.6\pct$         & $10.1\pct$     & 
& $5.1\pct$      & $5.5\pct$       
\\

\smallgap

Latent vector & PF           
& $92.9\pct$         & $40.4\pct$  &
& $88.1\pct$      & $13.2\pct$    
\\

Latent vector & Regression           
& $9.8\pct$   & $16.8\pct$   &  
& $5.6\pct$      & $7.3\pct$       
\\

\multicolumn{2}{l}{\smallgap \emph{Closest context}}          
& $11.0\pct$    & $39.3\pct$    &
& $11.4\pct$      & $34.5\pct$       
\\

\multicolumn{2}{l}{\smallgap \emph{Uninformed estimate}}             
& $1.7\pct$      & $3.3\pct$       &
& $2.9\pct$      & $3.7\pct$        
\\

\bottomrule
\end{tabular}
}
\vspace{-0.2cm}
\end{center}
\end{table}

\begin{table}[t]
\caption{Global localization results for the Street View dataset.}
\label{tab:main_street}
\begin{center}
\scalebox{.73}{
\begin{tabular}{llcc} 
\toprule

\multicolumn{2}{c}{Approach} & {Sequential } & {One-step } \\
Map & Algorithm &  loc. ($\uparrow$) & loc. ($\uparrow$) \\
\midrule

View-embed.& PF (ours)  
   & \best{73.2\pct} 
   & \best{28.4\pct} 
\\

View-embed. &  Regr. (ours)           
   & 42.3\pct{} 
      & $13.4\pct$   
\\
\hline

\smallgap

Latent image & PF                
    & $37.8\pct$ 
      & $9.8\pct$ 
\\

Latent image & Regression  
      & $11.1\pct$ 
      & $10.1\pct$  
\\

\smallgap

Latent vector & PF           
    & $15.8\pct$ 
     & $5.7\pct$  
\\

Latent vector & Regression           
 &  $10.4\pct$  
      & $7.1\pct$   
\\

\multicolumn{2}{l}{\smallgap \emph{Closest context}}          
   & $8.6\pct$ 
      & $8.7\pct$  
\\

\multicolumn{2}{l}{\smallgap \emph{Uninformed estimate}}             
     & $2.7\pct$  
        & $7.1\pct$  
\\

\bottomrule
\end{tabular}
}
\end{center}
\vspace{-0.2cm}
\end{table}

\begin{table}[t]
\caption{Tracking results for all datasets}
\label{tab:main_tracking}
\begin{center}
\scalebox{.73}{
\begin{tabular}{llccc} 
\toprule
\multicolumn{2}{c}{Approach} & 
\multicolumn{3}{c}{Tracking RMSE in m ($\downarrow$)}
\\
Latent map & Algorithm 
& Rooms & Mazes & Street View \\
\midrule

View-embed.  & DMN-PF (ours)  
& 2.76         &  \best{3.22}     & \best{4.15} 
\\

View-embed.  &  Regr. (ours)          
& \best{1.26}       & 4.52        & 6.39
\\
\hline

\smallgap

Image & PF                
& 2.90     & 3.41    & 6.13 
\\

Image & Regression  
& 9.06        & 9.06       & 9.51 
\\

\smallgap

Vector & PF           
& 2.97   & 6.07        & 7.75 
\\

Vector & Regression           
& 8.39       & 7.92         & 9.49   
\\

\multicolumn{2}{l}{\smallgap \emph{Closest context}}          
& 12.84         & 9.92        & 11.17
\\

\multicolumn{2}{l}{\smallgap \emph{Uninformed estimate}}             
& 14.75          & 13.19           & 13.15
\\

\bottomrule
\end{tabular}
}
\end{center}
\vspace{-0.2cm}
\end{table}

\subsection{Baselines}
\label{sec:alternatives}

We compare DMNs to the following baselines: unstructured latent map representations, (\emph{latent image} and \emph{latent vector}) similar to prior work~\cite{eslami2018neural, rosenbaum2018learning}; regression instead of particle filtering; an upper bound on image similarity based approaches (\emph{closest context}); and an \emph{uninformed estimate}.

\myparagraph{Latent image map.} %
We consider a representation similar to~\cite{eslami2018neural, rosenbaum2018learning}, which encodes the map as a $\dense H\mytimes W \mytimes D$ latent image without using explicit spatial structure. The latent image map can be also thought of as a 2D grid where each cell holds a learned latent vector. 
For a fair comparison, we extract features from context images using the same CNN as for view-embeddings, and average context features element-wise to obtain the latent image map. When using the map for localization we use the same network as in the DMN, which concatenates map and query image features to estimate particle log-likelihoods. Unlike in DMN, 
the context and particle poses enter the model by concatenating the coordinates with the corresponding context and query image features, respectively.
The number of learnable parameters is approximately the same as for view-embedding maps.

\begin{figure*}[!t]
\begin{floatrow}
\ffigbox[.13\textwidth]{%
  \centering
  \includegraphics[width=0.13\textwidth]{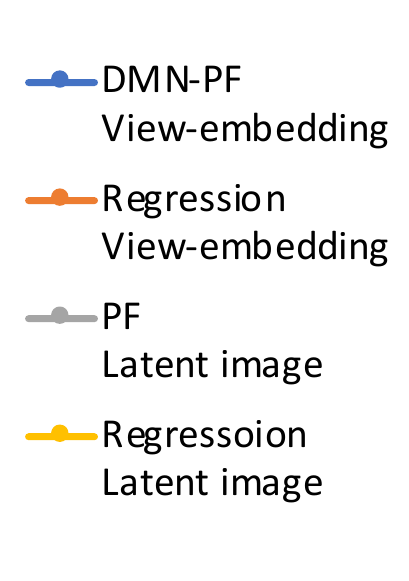}\vspace{.3cm}%
}{}
\ffigbox[.25\textwidth]{%
  \centering
  \includegraphics[width=0.25\textwidth]{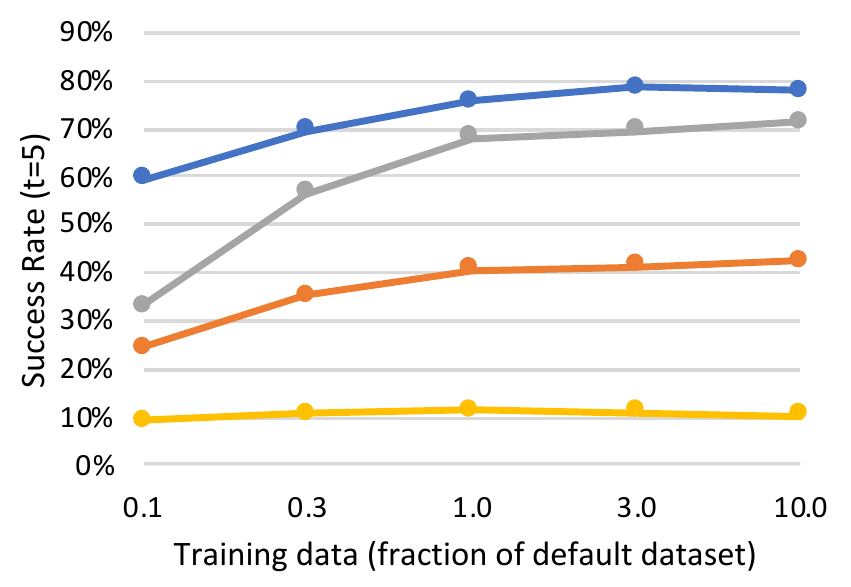}
}{\caption{Success as a function of the amount of training data}%
\label{fig:learning_curve}}
\hspace{.4cm}
\ffigbox[.25\textwidth]{%
  \centering
  \includegraphics[width=0.25\textwidth]{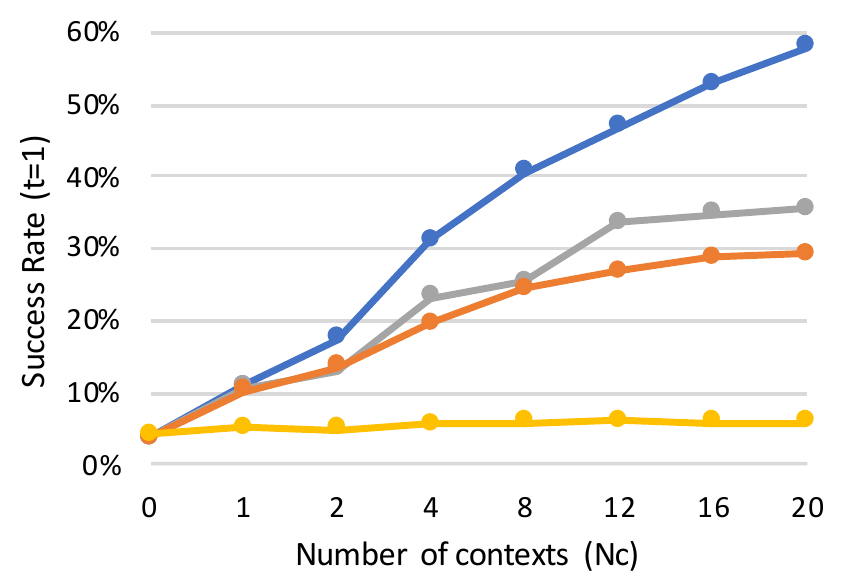}%
}{\caption{Train and test for different number of contexts $\dense N_c$.}%
\label{fig:fixed_context}
}
\ffigbox[.25\textwidth]{%
  \centering
  \includegraphics[width=0.25\textwidth]{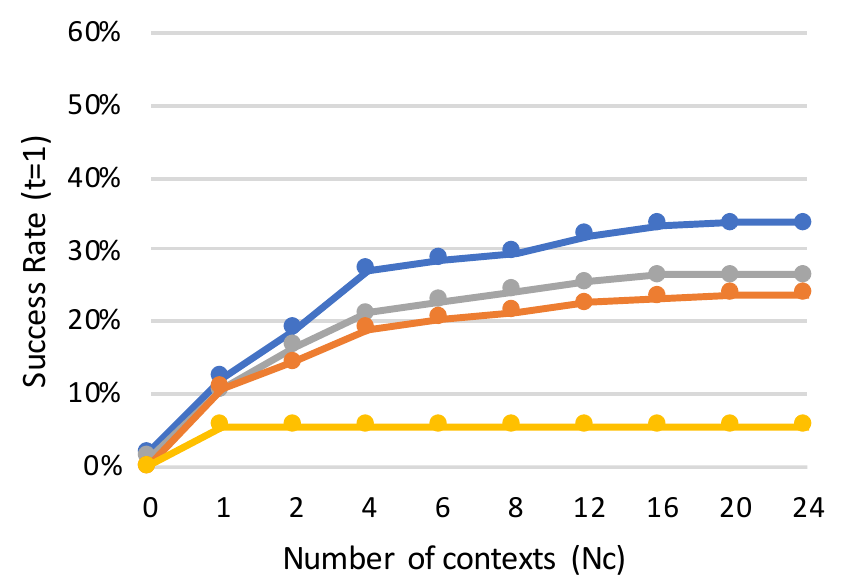}
}{\caption{Train with a fixed number of contexts $\dense N_c=4$, test with different $N_c$. }\label{fig:cross_context}}
\end{floatrow}
\end{figure*}

\myparagraph{Latent vector map.}
The latent vector map is similar to the latent image map, except that it extracts feature vectors using convolutional and dense layers. We selected hidden unit sizes to yield comparable number of parameters.

\myparagraph{Regression.}
We also compare with a regression network that directly regresses to a pose instead of performing particle filtering. The network structure is equivalent to a DMN with one particle. However, the particle pose is now set to the pose estimate of the previous step (or the mean initial belief for $\dense t=0$); and instead of estimating the particle likelihood, the observation model directly regresses to a pose ($x$, $y$, $\yaw{}$), which is interpreted relative to the previous pose estimate. The regression model has a similar number of learnable parameters as a DMN. %

\myparagraph{Closest context.} 
We report evaluation metrics for the context pose closest to the ground-truth query pose, which provides an upper bound on the performance of any image similarity-based approach.

\myparagraph{Uninformed estimate.} We report an uninformed estimate, the mean of the initial belief updated only with egomotion. The uninformed estimate helps calibrating the task difficulty.

\subsection{Uncertainty settings}
We consider two uncertainty settings, \emph{global localization} and \emph{tracking}, similar to~\cite{karkus2018particle}.
For global localization the initial belief is uninformed uniform over the environment, including orientation. 
For tracking the initial belief is a Gaussian around the first query pose with $\dense \sigma_x  = 6\meters, \sigma_y = 6\meters, \sigma_\yaw = 30^{\circ}$ and a mean offset sampled from the same distribution.

\subsection{Training and evaluation}
\label{sec:training_and_evaluation}

We train all models for one-step tracking with $\dense N_c=4$ contexts and $\dense K=32$ particles (when applicable). We terminate training if the validation error does not improve for 100k training iterations. Validation error is computed on 1000 samples every 20k iterations.  We use weight decay only for the Street View domain, with a scaler of $10^{-4}$. We performed a simple grid search over learning rates for each model and dataset independently, and picked the best performing parameter based on validation error.
The models are implemented in Tensorflow~\cite{tensorflow2015-whitepaper} and trained with the Adam optimizer~\cite{kingma2014adam}. Training took up to few days on an NVIDIA Tesla P100 GPU.

We evaluate trained models in multiple settings: one-step global localization ($\dense t=1$), sequential global localization ($\dense t=5$), and tracking ($\dense t=5$). During evaluation we increase the number of particles for PF: $\dense K=256$ for tracking and $\dense K=2048$ for global localization. The evaluation metric for tracking is RMSE in meters, \ie, the square root of the average MSE for the $x$--$y$ coordinates at time $t$. 
The evaluation metric for global localization is success rate, \ie, the percentage of trials where RMSE is under $8.94\meters$ at time $t$. This corresponds to 15.7\% of the environment area\footnote{The choice of MSE $\dense < 80 \approx 8.94^2$ was made in early experiments. Other thresholds would be also possible.}. %
All results are averages over 3 models trained with different seeds. We omitted standard errors which were small and would not have affected our conclusions.

\section{Results}

\subsection{Main results}
\label{sec:main_results}

The main results are presented in Tables~\ref{tab:main_joint}, \ref{tab:main_street}, \ref{tab:main_tracking}, encompassing three datasets, two uncertainty settings, single-step and sequential localization, and comparisons with alternative latent map representations and alternative localization approaches. Qualitative examples are shown in 
Figs.~\ref{fig:gqn_examples} and~\ref{fig:sv_example}.

We first consider global localization and present results for simulated environments in Table~\ref{tab:main_joint}. %
We observe that view-embedding maps perform consistently better than alternative map representations and particle filtering generally performs better than regression. While regression works well in the simple Rooms domain, we see a large benefit for DMN-PF in the more complex Mazes domain, (75.9\pct{} vs 39.9\pct{} at $\dense t=5$). PF can propagate a multimodal belief distribution over time, which is important to deal with ambiguous observations, while regression only propagates a point estimate\footnote{A regression network could output a parameterized distribution, e.g., a mixture of Gaussians, but propagating the distribution to the next step and handling non-standard multimodal distributions are non-trivial.}. The example in the last row of \figref{fig:gqn_examples} shows a PF belief with four distinct modes. 
Finally, the comparison with the \emph{closest context} shows that image-similarity based approaches cannot perform well in these domains.

Table~\ref{tab:main_street} shows global localization results for the Street View dataset. %
This is challenging real-world dataset. Baselines perform poorly and generally achieve less than 16\pct{} or 11\pct{} success. Sequential localization with PF is the only baseline with more reasonable success rates of 37.8\pct{}. In contrast, our method is able to achieve about 2x higher accuracy of 73.2\pct{}. This experiment shows more distinctly the importance of using view-based learned map (73.2\pct{} vs 37.8\pct{}) as well as the importance of particle filtering (73.2\pct{} vs 42.3\pct{}). \figref{fig:sv_example} shows an example for successful localization. Starting from a uniform belief, the DMN is able to provide a reasonable probabilistic estimate having as little as one observation. As new observations are received the belief is updated and the variance is further reduced.

\begin{figure}%
\begin{floatrow}
\ffigbox[.5\textwidth]{%
  \centering
  \includegraphics[width=0.5\textwidth]{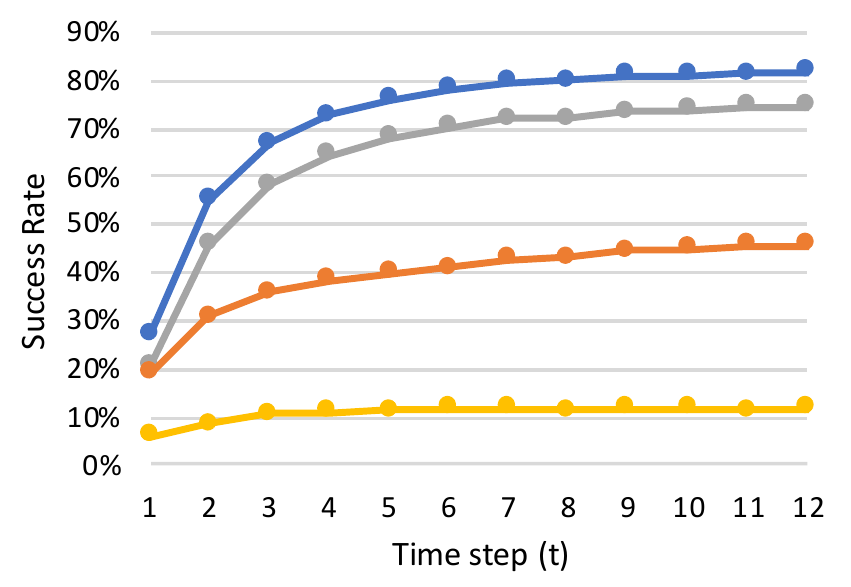}
  \vspace{-.4cm}
}{
\caption{Success as a func. of time.~~~~~}%
\label{fig:time_curve}
}
\hspace{-.5cm}
\ffigbox[.5\textwidth]{%
  \centering
  \includegraphics[width=0.5\textwidth]{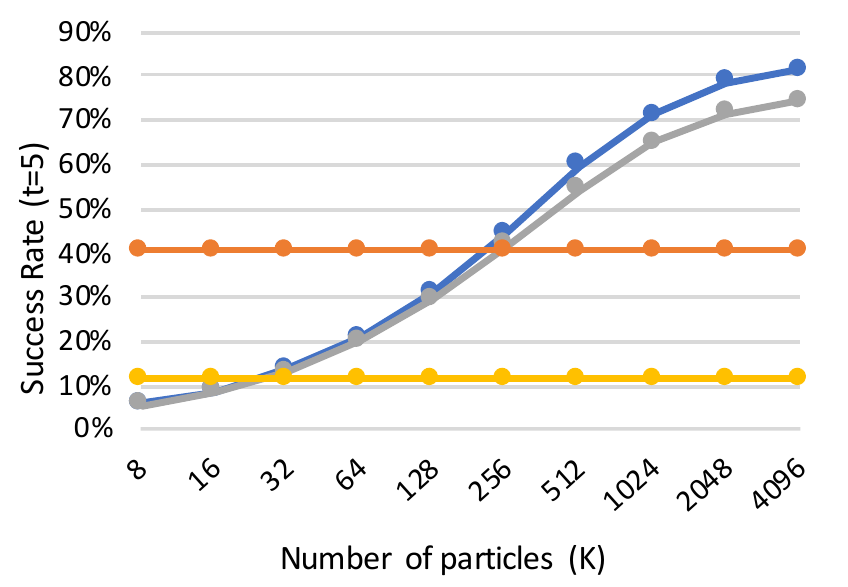}
  \vspace{-.4cm}
}{
\caption{Success as a function of the number of particles $K$.}
\label{fig:particles}}
\end{floatrow}
\end{figure}

\tabref{tab:main_tracking} evaluates models for sequential tracking at $\dense t=5$. We report RMSE for all datasets in meters. We observe similar tendencies as for global localization: view-embedding maps consistently perform better than alternatives, particle filtering is important in complex environments. The benefit of the DMN is the largest for the Street View dataset. %

\begin{figure*}[th]
\vspace{.09cm}%
\begin{floatrow}
\capbtabbox{
\scalebox{0.73}{%
\begin{tabular}{llcc} 
\toprule
\multicolumn{2}{c}{Approach} & 
\multicolumn{2}{c}{Tracking RMSE in m ($\downarrow$)}
\\
Map & Algorithm & Default env. & Large env.
\\
\midrule
View-embed. &  Regr. (ours)          
& \best{6.39}       & \best{7.01}      
\\
Latent image & Regr.  
& 9.51        & 32.77
\\
\bottomrule
\end{tabular}
}
\vspace{-0.2cm}
}{
\caption{Results for Large Street View environments%
}
\label{tab:large_environments}}%
\capbtabbox{
\scalebox{0.73}{%
\begin{tabular}{lc} 
\toprule
Map learning & Classification \\
objective &  accuracy ($\uparrow$) \\
\midrule
Localization       & 42.5\pct{} \\
Classification     & \best{98.5\pct{}} \\
\bottomrule
\end{tabular}
}
\vspace{-0.2cm}
}{
\caption{%
Global classification%
}\label{fig:global_classification}
}
\quad
\capbtabbox{
\scalebox{0.72}{%
\begin{tabular}{lc} 
\toprule
Approach  & Sequential loc. ($\uparrow$)  \\
\midrule
DMN-PF (ours)  & \best{75.9\pct} \\
without attention  & 73.0\pct{} \\
without egocentric  & 57.4\pct{} \\
without att. \& egoc.  & 37.8\pct{} \\
\bottomrule
\end{tabular}
}
\vspace{-0.2cm}
}{
\caption{Results for ablations of the DMN}
\label{tab:ablations}}
\end{floatrow}
\end{figure*}

\subsection{Analyzing the map representation}

\myparagraph{Data efficiency.} \figref{fig:learning_curve} compares models trained with different amounts of data for the Mazes domain. %
The benefit of view-embedding maps is even larger in the small data regime, which may be attributed to the spatial structure prior. %

\myparagraph{Number of contexts.} \figref{fig:fixed_context} and \figref{fig:cross_context} compare localization success as the number of contexts for mapping ($N_c$) changes in the Mazes domain. %
In \figref{fig:fixed_context} models are trained and evaluated with the same $N_c$ setting.  DMNs scale well with the number of contexts: performance monotonically improves, and the benefit of view-embedding maps increases with $N_c$.
In \figref{fig:cross_context} models are trained with a fixed $\dense N_c=4$, but evaluated with a range of $N_c$ settings. %
The performance is worse than for training in the same setting, as expected, but here too DMN with view-embedding maps generalize better.

\myparagraph{Larger environments.} %
We consider larger Street View environments for mapping, without changing the difficulty of localization. We start with $\dense N_c=4$ contexts as before, and add 4 new contexts that are located far from the originals. The initial belief remains the same. New contexts are irrelevant for localization because they are far from the original contexts (and the initial belief), but they make the environment to be mapped significantly larger.
\tabref{tab:large_environments} compares regression with different maps for tracking in default and large Street View environments. %
While view-embedding maps are almost unaffected (top row), the RMSE for unstructured image map increases dramatically (bottom row).
Latent image maps have a fixed capacity to represent the environment, and the map is built irrespective of the agent's current pose. In contrast, view-embedding maps have a flexible representation capacity that scales with the number of input contexts; and the egocentric spatial attention allows learning to attending only to views near to a candidate pose.

\subsection{Analyzing localization accuracy}

\myparagraph{Longer trajectories.} 
\figref{fig:time_curve} shows global localization results for longer trajectories in the Mazes domain, evaluated at different time steps $t$. 
DMN success rates increase rapidly until $\dense t=5$, and continue increasing monotonically up to over 80\pct. 
For regression, success rates increase slower and stay below 50\%. These results show the importance of propagating a probabilistic estimate with PF. %

\myparagraph{Number of particles.} 
\figref{fig:particles} shows sequential global localization results in the Mazes domain for a DMN trained with $\dense K=32$ particles and evaluated with a range of $K$ settings. Since regression networks have no particles, we plot constant performance for all $K$. DMN improves with more particles at test time, as one would expect from a standard PF. The comparison to regression highlights important benefits of PF, that is, 
higher accuracy and  
ability to trade off computation cost and performance at test time.

\subsection{Optimizing maps for the downstream task}
We analyze whether it is important to optimize the map representation for the downstream task.
We consider a new \emph{global classification} task, where the robot is given $\dense N_c=4$ context image-pose pairs, similarly to sparse localization, but each context is now taken from a different distant location. Given a new query image, the task is to choose the context closest to the unknown query pose.
For this experiment we tile multiple mazes together and sample context image-pose pairs from different mazes.  We use a model similar to the DMN, but with only $\dense K=N_c$ particles, where each particle pose is set to be a context pose. We treat observation log-likelihoods of DMN as class probabilities for classification over contexts, and train the model using a cross-entropy loss.

\tabref{fig:global_classification} compares modified DMN models where the map is optimized for either \emph{localization} or for \emph{classification}. The former trains a DMN for sparse localization, fixes the map representation, and further trains for global classification. The latter also trains a DMN for sparse localization, but then it adapts both the map representation and the observation model for global classification. Results show a significantly higher performance when optimizing the map representation for the downstream classification task.

\subsection{Ablations}
\tabref{tab:ablations} compares ablations of the DMN 
for sequential global localization in the Street View domain. We independently remove egocentric coordinate transformations and attention weighting. %
Results show that egocentric transformations are critical, and attention  provides a small additional improvement. The view-embedding map appears to be important in DMN, as it enables the egocentric transformations.

\section{Conclusion}
We introduced the differentiable mapping network that learns a map for sparse visual localization, and demonstrated strong performance across simulated and real-world domains. \blfootnote{\myparagraph{Acknowledgements.} We thank Dan Rosenbaum for fruitful discussions on the approach and for suggesting the Street View dataset. We thank Chad Richards for editing the manuscript.}
We believe these results provide useful insights for applications beyond the sparse localization domain.
Future work may explore extensions of our view-based latent map structure, e.g., having multiple view-embeddings at learned locations. 
A particularly interesting direction is to extend the proposed work to visual SLAM, 
where new observations could be treated as additional context, and uncertain context poses could be encoded in particles.

\clearpage

\balance
\bibliography{references}
\bibliographystyle{ieeetr}

\end{document}